\documentclass{article}

\usepackage[preprint]{neurips_2024}

\usepackage[utf8]{inputenc} 
\usepackage[T1]{fontenc}    
\usepackage{hyperref}       
\usepackage{url}            
\usepackage{booktabs}       
\usepackage{amsfonts}       
\usepackage{nicefrac}       
\usepackage{microtype}      
\usepackage{xcolor}         
\usepackage{amsmath}
\usepackage{graphicx}
\usepackage{caption}
\usepackage{blindtext}

 \usepackage{multirow}
 \usepackage{multicol}
 \usepackage{subcaption}
 \usepackage[export]{adjustbox}

\newcommand{\keypoint}[1]{\noindent\textbf{#1}\quad} 
\usepackage{natbib}
\hypersetup{
  colorlinks   = true, 
  urlcolor     = blue, 
  linkcolor    = red, 
  citecolor   = blue 
}

\title{Memorized Images in Diffusion Models share \\a Subspace that can be Located and Deleted}

\author{Ruchika Chavhan$^1$, Ondrej Bohdal$^1$, Yongshuo Zong$^1$,  Da Li$^2$, Timothy Hospedales$^{1,2}$\\
$^1$University of Edinburgh, \\
$^2$Samsung AI Research Centre, Cambridge\\
}

\begin{document}

\maketitle

\begin{abstract}
Large-scale text-to-image diffusion models excel in generating high-quality images from textual inputs, yet concerns arise as research indicates their tendency to memorize and replicate training data, raising We also addressed the issue of memorization in diffusion models, where models tend to replicate exact training samples raising copyright infringement and privacy issues. Efforts within the text-to-image community to address memorization explore causes such as data duplication, replicated captions, or trigger tokens, proposing per-prompt inference-time or training-time mitigation strategies. In this paper, we focus on the feed-forward layers and begin by contrasting neuron activations of a set of memorized and non-memorized prompts. Experiments reveal a surprising finding: many different sets of memorized prompts significantly activate a common subspace in the model, demonstrating, for the first time, that memorization in the diffusion models lies in a special subspace. Subsequently, we introduce a novel post-hoc method for editing pre-trained models, whereby memorization is mitigated through the straightforward pruning of weights in specialized subspaces, avoiding the need to disrupt the training or inference process as seen in prior research. Finally, we demonstrate the robustness of the pruned model against training data extraction attacks, thereby unveiling new avenues for a practical and one-for-all solution to memorization. Our code is available at \url{https://github.com/ruchikachavhan/editing-memorization}.

\end{abstract}

\section{Introduction}

Recent advancements in diffusion models (DMs) have showcased remarkable capabilities in image generation. Particularly, text-to-image (T2I) diffusion models such as DALL-E and Stable Diffusion \citep{luccioni2023stable} excel in creating high-quality images that accurately correspond to textual prompts. However, growing research \citep{somepalli2022diffusion,carlini2023extracting} suggests that these models can memorize their training data, as some seemingly ``novel'' creations are almost identical to images within their training datasets.
This memorization issue raises significant concerns regarding copyright infringement of the original training data and heightens the risk of leaking privacy-sensitive information, causing immense legal troubles in privacy-critical fields like medical imaging or finance.

Memorization is increasingly often being addressed in discriminative models \citep{NEURIPS2020ea89621b, NEURIPS2022564b5f82, shokri2017membership, tramer2022truth} and pre-trained language models \citep{petroni2019language, carlini2023quantifying, hartmann2023sok}. However, ongoing debate about the cause of memorization still persists. Some argue that memorization is a prerequisite for generalization, as models tend to generalize well despite frequently overfitting the data -- a phenomenon often referred to as \textit{benign overfitting}. Despite its prevalence in T2I generation, this issue is understudied and poorly documented as the cause of memorization in DMs remains unclear, with varying opinions across different studies.

Recent research on memorization in diffusion models \citep{wen2024detecting, ren2024unveiling, yoon2023diffusion, gu2024on, chen2024memorizationfree, somepalli2023understanding} attributes this phenomenon to data duplication and the presence of highly specific text prompts in training data that \textit{trigger} memorization. Specifically, \citet{wen2024detecting, ren2024unveiling, somepalli2023understanding} demonstrate that for such memorized prompts, the text consistently steers the generation towards memorized solutions, irrespective of initial conditions. Subsequently, they introduce mitigation strategies that include inference-time techniques, such as detecting and perturbing \textit{trigger} tokens, and training-based methods such as filtering training data to reduce duplications and perturbing the training data. Nevertheless, current memorization mitigation strategies interfere either with the training or the inference pipeline of diffusion models.

In this paper, we present a surprising observation that memorization can be localized within a distinct and narrow subset of neurons of pre-trained diffusion models. Diverging from prior research that pinpoints memorization on a per-prompt level, we identify there are critical neurons within pre-trained models that exhibit heightened responses for a small subset of memorized prompts, compared to non-memorized prompts. We coin the term \textit{memorized neurons} to represent these neurons. More interestingly, the set of \textit{memorized neurons} identified for different subsets of memorized prompts are highly overlapped, suggesting, for the first time, that memorization lies within a specialized subspace in pre-trained diffusion models.

We leverage this discovery to develop a one-time training-free strategy for addressing the issue of memorization in diffusion models. Our approach involves posthoc surgery, wherein we selectively prune regions in weight space that act on these memorized neurons. Unlike traditional memorization mitigation techniques, our method offers a significant advantage in terms of ease and speed, as it does not necessitate modifications to the training or inference processes of diffusion models. Furthermore, we showcase the robustness of the pruned model against training data extraction attacks, thereby unveiling promising avenues for a practical and comprehensive solution to memorization.


\section{Related Work}

\textbf{Memorization in diffusion models.} Membership inference attacks \citep{webster2023reproducible, carlini2023extracting} demonstrate that memorization in DMs can be categorized into three main types: 1)\textit{ matching verbatim}: where the images produced from the memorized prompt are an exact pixel-for-pixel match with the original training image; 2) \textit{retrieval verbatim}: where the generated images perfectly correspond to some training images but are paired with different prompts; 3) \textit{template verbatim}: where the generated images partially resemble training images, though there may be variations in colors or styles.

Recent research delves into the causes of memorization in DMs, attributing the phenomenon to factors such as image duplication \citep{somepalli2022diffusion, gu2024on}, the presence of highly specific tokens in text prompts that \textit{trigger} memorization \citep{somepalli2023understanding, wen2024detecting, ren2024unveiling}, and an excessive number of training steps that lead to overfitting on a subset of samples which the model fails to generalize on \citep{somepalli2022diffusion}. Based on these observations, studies have identified markers of memorization, such as a disproportionate focus on specific tokens in cross-attention \citep{ren2024unveiling} and higher magnitude of text-conditional predictions \citep{wen2024detecting}, which are then utilized for detecting memorized prompts and trigger tokens. Subsequently, these works \citep{wen2024detecting, somepalli2022diffusion, ren2024unveiling} introduce two mitigation pipelines: inference-time, where trigger tokens are perturbed, and training-time, where the model is fine-tuned by training on identified non-memorized subsets.

However, training-time mitigation strategies can be ineffective as prior research \citep{carlini2022privacy} demonstrates an \textit{onion-peel effect} of memorization, wherein excluding memorized samples from training does not mitigate memorization, rather it reveals a new ``layer'' of previously private points that are now memorized by the model. Moreover, this phenomenon has not been highlighted in previous works as they only evaluate on memorized samples excluded from fine-tuning and do not consider new samples that the model might have memorized. Additionally, the inference time mitigation strategies introduce an additional step in the pipeline which requires formulation of heuristics to detect and perturb triggering text tokens.

Unlike previous approaches that address memorization on a per-prompt basis, our study seeks to \textit{localize} memorization within off-the-shelf pre-trained models and subsequently edit the model by eliminating the regions critical for memorization, thus introducing a one-time, training-free strategy.

\textbf{Localising memorization in classification models.} Previous research \citep{maini2023neural} in discriminative models indicates that the memorization of particular ``hard'' or outlier training samples tends to be concentrated in a few neurons or convolutional channels scattered across different layers of the model. They also demonstrate that excluding these neurons during test time effectively mitigates memorization without compromising the original model's performance. In contrast to methods outlined in \citep{maini2023neural}, which necessitate costly gradient calculations and monitoring of heuristics during training from scratch, our approach operates exclusively on pre-trained models.  Moreover, to the best of our knowledge, our work is the first to explore this premise in the domain of diffusion models.

The subsequent sections are structured as follows: Sections \ref{sec:background}  provide an overview of diffusion models and delves into the phenomenon of memorization, laying the groundwork for our approach. In Section \ref{sec:mem-method}, we outline our method for identifying neurons in pre-trained diffusion models that exhibit heightened receptivity to a small subset of memorized prompts compared to non-memorized ones. Following this, we present a surprising observation in Section \ref{sec:mem-localisation}: neurons indicative of various memorized subsets share high similarities, suggesting that memorization can be localized to specific regions within pre-trained models. Subsequently,  eliminating these neurons effectively \textit{edits} memorization without the need for retraining.


\section{Background}
\label{sec:background}
\paragraph{Diffusion models.}

Diffusion models (DMs) are trained to denoise images by reversing a forward Markov process, where noise is incrementally added to input images over several time steps $t \in [0, T]$. During the training phase, given an original image $\mathbf{x}_0$, a noisy version of the image $\mathbf{x}_t$ at time $t$ is generated using $\sqrt{\alpha_t} \mathbf{x}_0 + \sqrt{1-\alpha_t} \varepsilon$, where $\varepsilon \sim \mathcal{N}(0, I)$ and $\alpha_t$ is a parameter that decreases over time. The model learns to estimate the noise added to obtain $\mathbf{x}_t$ so that the original image $\mathbf{x}_0$ can be recovered by removing the noise from $\mathbf{x}_t$.

In this paper, we primarily focus on Latent Diffusion Models (LDMs) \citep{rombach2021highresolution}, which offer a significant advantage by speeding up the forward and reverse diffusion process by operating in the latent space of the input $\mathbf{x}$, represented as $\mathbf{z}$. Typically, image encoders like CLIP \citep{radford2021learning} are used to extract the latent $\mathbf{z}_0$ from real image $\mathbf{x}_0$, and a VAE decoder maps the latent space back to images. Thus, a LDM consists of a latent embedding denoiser $\epsilon_\theta(.)$, which is trained to predict the added noise by stochastically minimizing the objective $\mathcal{L}(\mathbf{z}, p)=\mathbb{E}_{\varepsilon, \mathbf{x}, p, t}\left[\left\|\varepsilon-\epsilon_\theta\left(\mathbf{z}_t, t, p\right)\right\|\right]$ given a text prompt $p$.

Text-conditional diffusion models, such as Stable Diffusion, employ classifier-free diffusion guidance \citep{rombach2021highresolution} to steer the sampling process toward the desired condition. This is achieved by combining the conditional and unconditional predictions, as shown in Equation \ref{eq:cls-free}, enabling the model to effectively guide itself:
\begin{equation}
\label{eq:cls-free}
    \hat{\epsilon}_\theta\left(z_t, t, p\right) \leftarrow \epsilon_\theta\left(z_t, t\right)+s \cdot\left(\epsilon_\theta\left(z_t, t, p\right)-\epsilon_\theta\left(z_t, t\right)\right)
\end{equation}

\paragraph{A close look at memorization in DMs.}


The prevailing understanding is that a memorized image can be reproduced from the training data regardless of the random initialization of the latent space \citep{ren2024unveiling, wen2024detecting, somepalli2022diffusion, somepalli2023understanding}. A simple look at classifier-free guidance in Equation \ref{eq:cls-free} suggests that if $|\epsilon_\theta(z_t, t, p)| \gg |\epsilon_\theta(z_t, t)|$ with a reasonable scaling value $s$, then the text-conditional term starts to heavily dominate the combined prediction. This has also been demonstrated in \citep{wen2024detecting}, which discovers that for memorised prompts the value of $\sum_{t=1}^T\|\left(\epsilon_\theta\left(z_t, t, p\right)-\epsilon_\theta\left(z_t, t\right)\right)\|_2$ is significantly higher than the one for non-memorised prompts.

{Building upon this insight, our approach first identifies neurons that exhibit significantly higher activation levels for conditional predictions associated with the memorized subset \( P \), in contrast to unconditional predictions derived from passing a null string \( p_\phi \) through the model.}

\section{Methodology}

\label{sec:mem-method}

{Recent papers in the domain of Large Language models (LLMs) have proven the existence of certain neurons that specialize in different functions \citep{zhang2023emergent, suau2020finding} and are critical for safety responses \citep{wei2024assessing}. They draw inspiration from pruning \textit{expert} modules in the network \citep{zhang2022moefication, zhang2023emergent} and utilize pruning techniques \citep{sun2024simple, lee2019snip} to determine a set of neurons critical to the safety of LLMs. In line with their spirit, we propose to localize certain neurons in DMs for memorization issues and prune them to address the defect. To this end, we repurpose a recent pruning approach, Wanda \citep{sun2024simple}, to discover and prune memorization neurons of DMs.}


\textbf{Wanda pruning~\citep{sun2024simple}:}  We begin by denoting the weights of a linear layer by $\mathbf{W} \in \mathbb{R}^{d^\prime \times d}$ and inputs $\mathbf{Z} \in \mathbb{R}^{d \times n}$, where $n$ is the number of samples. 
\citet{sun2024simple} estimates the collective impact of both weights and feature magnitudes on neuron activations, enabling the exploration of important neurons (from weights) for specific concepts (from input features). As a result, the importance score of each element of the weight matrix is given by an element-wise product of its magnitude and the $\ell_2$ norm of corresponding input features. Specifically, the score of a weight given an input is computed as:
\begin{equation} 
\label{eq:wanda}
    \mathbf{S}_{(i,j)} =\left|\mathbf{W}\right|_{(i,j)} \cdot \left\|\mathbf{Z}_{(j,:)}\right\|_2,
\end{equation}
where $|\cdot|$ computes the absolute value, and $\| \cdot \|_2$ denotes the $l_2$-norm. For the $i$-th row of $\mathbf{W}$, the bottom $s$\% weights with the lowest scores among $\mathbf{S}_{(i,:)}$ are zeroed out in \citet{sun2024simple}, which can be referred to for more details.

\textbf{Candidate neurons to prune in DMs:} Image denoisers in popular LDMs, such as Stable Diffusion, are characterized by the use of UNets \citep{ronneberger2015unet}. UNets consist of ResNet blocks that downsample or upsample the denoised latent space representations and transformer blocks that consist of self-attention between latent space, cross attention to incorporate textual guidance, and a Feed-forward network (FFN) with GEGLU activation function \citep{shazeer2020glu}. This paper focuses on weight neurons in these two-layer feed-forward networks, specifically its \emph{second} linear layer.

At time step $t$ and layer $l$, we denote the input to the FFN for text prompt $p$ by $z^{t,l}(p) \in \mathbb{R}^{d \times m}$ and output of the FFN by $z^{t,l+1}(p) \in \mathbb{R}^{d \times m}$. Here $m$ is the number of latent tokens. FFN in Stable Diffusion consists of GEGLU activation \citep{shazeer2020glu}, which operates as shown in Equation \ref{eq:geglu}:
\begin{gather}
\label{eq:geglu}
h^{t,l}(p) = \operatorname{GEGLU}(\operatorname{Linear}(z^{t,l}(p)) \\ \nonumber
     z^{t,l+1}(p) = \mathbf{W}^l \cdot h^{t,l}(p),
\end{gather}
where $\mathbf{W}^l \in \mathbb{R}^{d \times d'}$ is the weight matrix in the \emph{second} linear layer.
Next, we outline our framework for identifying \textit{memorized neurons} in this linear layer within the FFN layers using the importance score described above.

\subsection{Localizing and Pruning Memorized Neurons}


\textbf{Layer-wise Wanda score for memorized prompts at time $t$:} 
Membership inference attacks \citep{webster2023reproducible, carlini2023extracting} have demonstrated that DMs generate exact training data \citep{schuhmann2022laionb} using a similarity metric between generated and training data images. We begin by randomly sampling a subset of $n$ memorized prompts out of 500 prompts discovered by the extraction attack introduced in \citet{webster2023reproducible}. We denote this set of memorized prompts by $P = \{p_1, p_2, ... p_n\}$.

We collect neuron activations corresponding to the set of known memorized prompts $P$ and arrange them in a matrix denoted by $\mathbf{H}^{t,l}(P) = [h^{t,l}(p_1), h^{t,l}(p_2), ..., h^{t,l}(p_n)]$ such that $\mathbf{H}^{t,l}(P) \in \mathbf{R}^{d^\prime \times n}$. Note that this process only requires one forward pass per prompt. Then, we calculate the importance score for FFN weights $\mathbf{W}^l$ using input neurons for memorized prompts using Equation \ref{eq:wanda} as:
\begin{equation}
    \label{eq:wanda-sd}
    \mathbf{S}^{t,l}(P)_{(i,j)} = \left|\mathbf{W}^l\right|_{(i,j)} \cdot \left\|\mathbf{H}^{t,l}(P)_{(j,:)}\right\|_2 
\end{equation}
Similarly, we calculate the importance score for the null prompt $p_\emptyset$ as $\mathbf{S}^{t,l}(P_\emptyset)$, where $P_\emptyset$ is formulated by stacking $n$ repetitions of $h^{t,l}(p_\emptyset)$. 

\textbf{Localizing and pruning memorized neurons:} Similar to \citet{wei2024assessing}, we collect the indices of the important neurons considering the highest Wanda scores in each row of the weight matrix. Specifically, for a given sparsity level $s$\%, we define the top-$s$\% important neurons in the $i$-th row of $\mathbf{W}^l$ as
\begin{equation}
    \mathbf{A}^{t, l}({P}) = \{(i,j) | \ \ \text{if} \  \ \mathbf{S}^{t,l}({P})_{(i, j)} \ \ \text{in} \ \  \operatorname{top-s\%} (\mathbf{S}^{t,l}({P})_{(i, :)}) \}.
\end{equation}
Intuitively, $\mathbf{A}^{t, l}({P})$ denotes the set of weight neurons that offers the highest contribution to the denoised predictions in the reverse diffusion process at time step $t$ for the prompt set $P$. 

We now compare the Wanda scores of the most important weight neurons  $\mathbf{A}^{t, l}({P})$, with their importance scores when corresponding to the null string. A weight neuron is defined as a \textit{memorized} neuron if it ranks among the top $s$-\% of important neurons and its Wanda score exceeds that of the null string. We define the set of memorized neurons denoted by $\mathbf{V}^{t,l}(P, P_\emptyset)$ which is formulated as  
\begin{equation}
    \mathbf{V}^{t,l}(P, P_\emptyset) = \{(i,j) | \ \ \text{if} \  \ \mathbf{S}^{t,l}({P})_{(i, j)} >  \mathbf{S}^{t,l}(P_\emptyset)_{(i, j)} \ \ \ \forall (i, j) \in \mathbf{A}^{t, l}(P)\}
\end{equation}

To prune the memorized neurons, we first aggregate the indices across different time steps and zero out a weight neuron if its index is in $\mathbf{V}^{t, l}({P, P_\emptyset})$.
\begin{equation}
    \mathbf{W}^l_{(i,j)} = 0 \ \ \text{if} \ \  (i,j) \in \underset{t={T, T-1, ..., T - \tau}}{\cup} \mathbf{V}^{t, l}({P, P_\emptyset}),
    \label{eq:mask-union}
\end{equation}
then we will use the pruned $\mathbf{W}^l$ for image sampling mitigating the prompt memorization. Empirically, we find that aggregating a small number $\tau$ of time steps is enough for memorization mitigation and quality image generation.

\section{Memorization can be Localized and Edited within a Small Subspace}
\label{sec:mem-localisation}
\subsection{Memorized Neurons can be Localized within a Small Subspace}


\textbf{Experimental setup.} To evaluate our method, we use 500 memorized prompts identified for Stable Diffusion v1 \citep{webster2023reproducible} and denote this dataset by $\mathcal{D}$. We select $N$ different subsets of prompts from $\mathcal{D}$, each containing $m$ memorized prompts. We denote the collection of these subsets by $\mathbb{P}^{N, m} = \{P^i\} \ \forall i \in [1, N]$, such that $|P^i| = m$. In the rest of this paper, we use the term \textit{collection} to denote $\mathbb{P}^{N, m}$ and \textit{subset} to denote $P^i$ for $i \in [1, N]$.

We utilize Stable Diffusion v1.5, which consists of 16 FFN layers, denoted by $L$. During inference, noisy images are sampled with a fixed random seed and denoised over 50 iterations. As per Section~\ref{sec:mem-method}, for a subset $P^k \in \mathbb{P}^{N,m}$, we first collect the activations of all prompts $\mathbf{H}^{t,l}(P^k)$ to obtain the importance score $\mathbf{S}^{t,l}(P^k)$ for weights of layer $l$ at time step $t$ using Equation \ref{eq:wanda-sd}. Subsequently, memorized weight neurons  in $W_l^2$ are discovered by formulating the set $\mathbf{V}^{t, l}(P^k)$ as shown in Equation \ref{eq:mask-union}. In the following experiment, we use a sparsity threshold of $s = 1\%$.

Now, we methodically demonstrate that memorized neurons discovered from different subsets of memorized prompts in $\mathbb{P}^{N, m}$ are highly similar, indicating that memorized prompts activate a common subspace in the weight space of pre-trained models. In this section, we present our analysis along two dimensions: denoising time steps and layers. This allows us to visualize the similarities in memorized neurons across the denoising trajectory and throughout the depth of the diffusion model.

\begin{figure}
    \centering
    \includegraphics[width=0.85\linewidth]{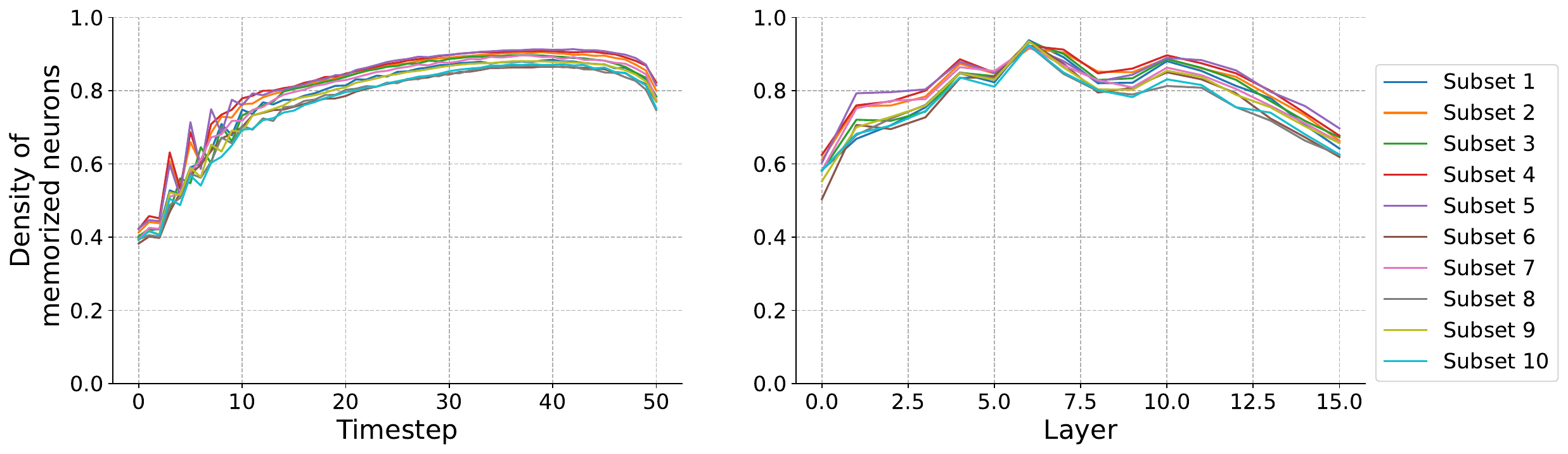}
    \caption{Density of memorized neurons averaged over timestep (left) and layer (right) for 10 different subsets containing 10 prompts each. We observe that the number of neurons identified as \textit{memorized} is similar across different subsets.}
    \label{fig:density}
\end{figure}

\keypoint{Different subsets yield a comparable number of memorized neurons.}
We define the \textit{density} of memorized neurons, denoted by $d^{t, l}(P^k)$, as the percentage of elements in the time-dependent set of memorized neurons $V^{t, l}(P^k)$ in Equation \ref{eq:mask-union}. Our objective is to compare the density of memorized neurons discovered from different subsets across the denoising steps and layers. Therefore, we calculate the densities averaged over time $d^l(P^k) = \sum_{t=0}^T d^{t, l}(P^k)$ and average over layer $d^t(P^k) = \sum_{l=0}^{L} d^{t, l}(P^k)$. In Figure \ref{fig:density}, we present the average densities $d^l(P^k)$ and $d^t(P^k)$ for all $P^k \in \mathbb{P}^{N, m}$. In this experiment, we consider $N=10$ and $m=10$. First of all, we observe that all subsets activate a very compact set of neurons, as indicated by densities less than 1\%. Our initial intriguing discovery is the striking similarity in the number of memorized neurons found across different subsets.

\begin{figure}
    \centering
    \includegraphics[width=0.75\linewidth]{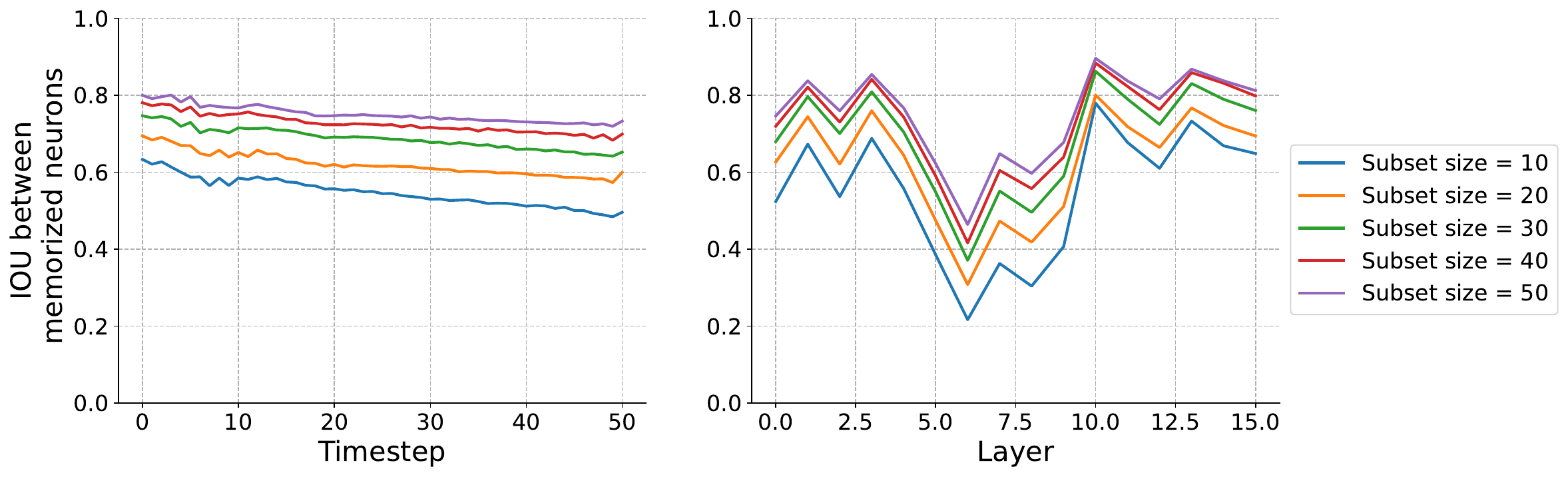}
    \caption{Average Pairwise IOU averaged over timestep (left) and layer (right) for $N=10$ and varying subset sized $m$.}
    \label{fig:iou-different-subsets}
\end{figure}

\textbf{The sets of memorized neurons for each memorized prompt set are very similar.} We proceed to compute the average pairwise intersection-over-union (IOU) for time step $t$ and layer $l$  between the memorized neurons activated by two distinct subsets within $\mathbb{P}^{N, m}$. Let us denote the function that calculates the IOU between two binary matrices $A$ and $B$ as $\text{iou}(A, B)$. We calculate the \textit{average pairwise} Intersection-Over-Union (IOU) for a collection $\mathbb{P}^{N,m}$ at a single time step $t$ and layer $l$ by $\text{IOU}^{t, l}(\mathbb{P}^{N,m})$. This is derived by averaging the IOU values between all pairs of subsets within $\mathbb{P}^{N,m}$, represented as $\text{IOU}(\mathbb{P}^{N,m}) = \frac{1}{n(n-1)}\sum_{i \neq j}^N \text{iou}(\mathbf{V}^{t, l}(P^i),\mathbf{V}^{t, l}(P^j))$.

Similar to previous visualizations of memorized neuron density, we compute the average pairwise IOU over time steps and layers, implemented as $\sum_{t=0}^T \text{IOU}^{t, l}(\mathbb{P}^{N,m})$ and $\sum_{l=0}^L \text{IOU}^{t, l}(\mathbb{P}^{N,m})$ respectively. We replicate this experiment across different collections, maintaining a fixed number of subsets $N$ at 10 and varying the size of each subset $m$ from 10 to 50. Figure \ref{fig:iou-different-subsets} illustrates two striking findings:
\begin{itemize}
    \item Figure \ref{fig:iou-different-subsets} (left) illustrates that within a single collection with fixed values of $N$ and $m$, the average IOU remains consistently high across all denoising iterations. This suggests that different subsets activate similar sets of memorized neurons along the denoising trajectory.
    \item  Figure \ref{fig:iou-different-subsets} (right) illustrates that in certain layers of the UNet, distinct subsets activate remarkably similar sets of memorized neurons. This phenomenon is particularly pronounced in the early down-sampling blocks and the up-sampling blocks of the UNet.
\end{itemize}

Our approach, which entails identifying a subset of memorized neurons for a given set of memorized prompts, reveals that discovered memorized neurons exhibit significant similarity across different subsets of memorized prompts. Subsequently, we demonstrate that mitigating memorization is indeed achieved by eliminating these memorized neurons through model pruning.

\begin{figure}
    \centering
        \begin{subfigure}[b]{0.4\textwidth}
        \centering
\includegraphics[width=\textwidth]{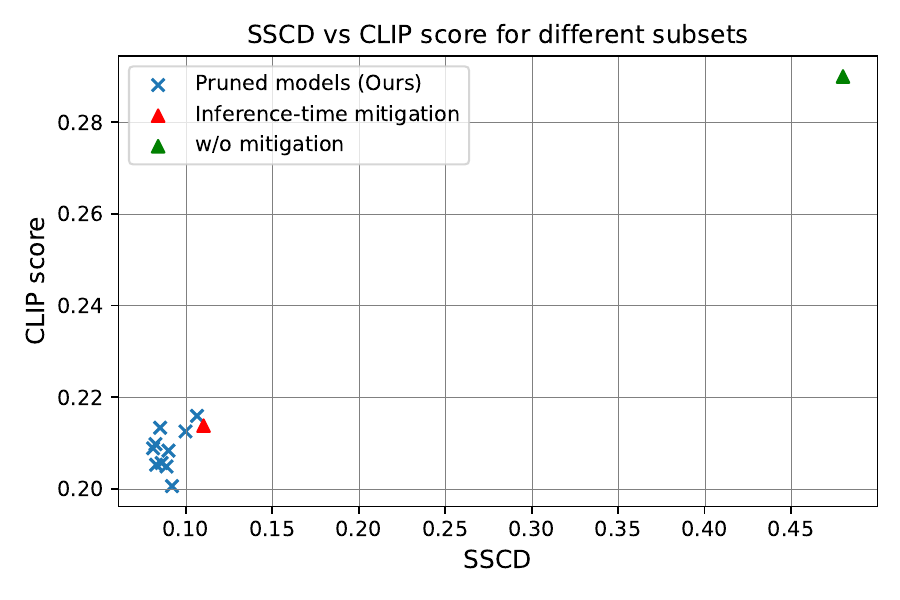}
    \end{subfigure}
    \begin{subfigure}[b]{0.5\textwidth}
        \centering \includegraphics[width=\textwidth]{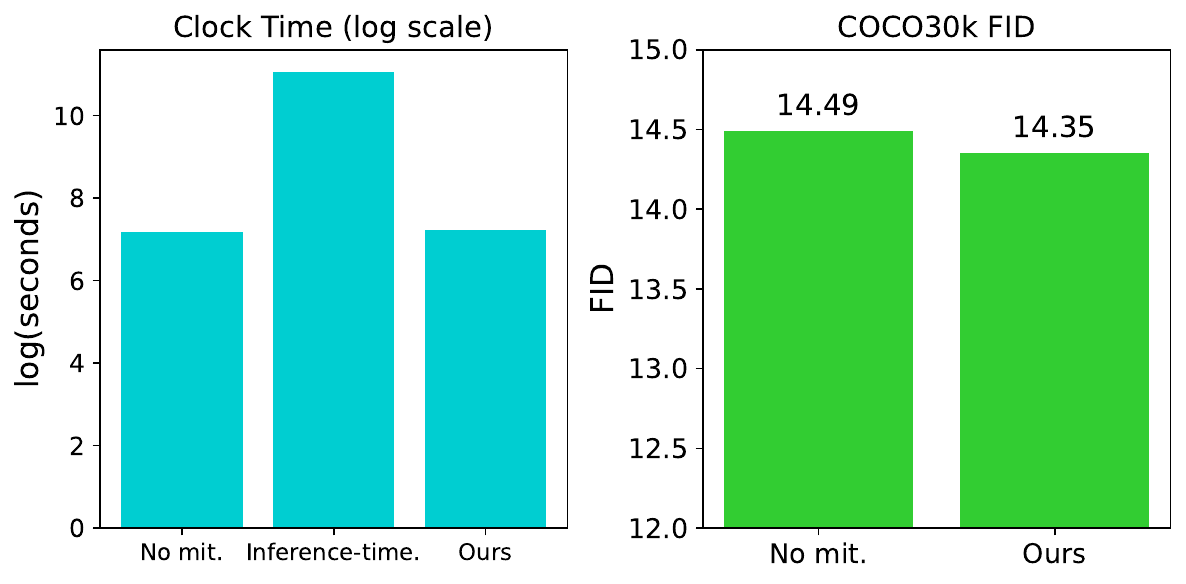}
        \label{fig:sub1}
    \end{subfigure}
    \caption{\textit{Left}: Quality  (CLIP similarity score, $\uparrow$) vs Memorization (SSCD, $\downarrow$) for 10 different pruned models compared with inference-time mitigation in \citet{wen2024detecting}. All the pruned models show less memorization than the no-mitigation baseline indicating that memorization can be edited via model pruning. \textit{Right}: Clock Time and COCO30k FID for baselines and our proposed approach. We provide similar generation quality and memorization reduction than \citet{wen2024detecting}, but substantially faster inference.}
    \label{fig:editing-memorization}
\end{figure}

\subsection{Memorized Images Can be Edited via Pruning Memorized Neurons}
\label{sec:mem-pruning}

{Starting with a collection $\mathbb{P}^{N,m}$, we initiate our experiment by pruning memorized neurons from a pre-trained Stable Diffusion model according to Equation \ref{eq:mask-union} for each subset $P^k \in \mathbb{P}^{N,m}$. The resulting pruned weights $\hat{W}^l$ substitute the pre-trained FFN weights, while the remainder of the model remains unchanged. We denote the pruned model obtained from utilizing memorized prompts in a subset $P^k$ as $\hat{\epsilon}_\theta(P^k)$. In this section, we fix $N=10$ and the size of the subsets $m=10$.}

As observed in Figure \ref{fig:density}, we alter an extremely compact subspace of approximately 1\% of the weights in the FFNs, regardless of the memorized subsets considered to obtain the pruned model. In this section, we illustrate that pruning the compact subspace substantially alleviates  memorization.

\keypoint{Evaluation setup.}
We evaluate the set of pruned models $\{\hat{\epsilon}_\theta(P^k);  P^k \in \mathbb{P}^n\}$ on the dataset of 500 memorized prompts $\mathcal{D}$ released by \citet{webster2023reproducible}. Note that subsets in $\mathbb{P}$ contain prompts that were sampled from $\mathcal{D}$. Therefore, for a fair comparison, to evaluate a model $\hat{\epsilon}_\theta(P^k)$, we remove all the prompts in $P^k$ from $\mathcal{D}$ to form the test sets.

\keypoint{Metrics and baselines.} We assess the extent of memorization by comparing the generated image with the original image, and the CLIP similarity score to quantify the alignment between the generated image and its corresponding prompt. Lower SSCD values indicate less memorization, while higher CLIP values indicate greater similarity to the text prompt. We additionally compare our editing method with two baselines: (1) Pre-trained Stable Diffusion (also referred to as No-mitigation in this section), and (2) Inference-time mitigation proposed in \citet{wen2024detecting}, which is based on token perturbation during inference.

\keypoint{Comparing our approach with baselines.} We present the CLIP Similarity vs SSCD for the set of pruned models $\{\hat{\epsilon}_\theta(P^k);  P^k \in \mathbb{P}^n\}$ in Figure \ref{fig:editing-memorization}. We observe that all pruned models exhibit decreased SSCD compared to the No-mitigation baseline and comparable SSCD to Inference-time mitigation in \citet{wen2024detecting}. However, it is important to note that inference-time mitigation methods \citep{wen2024detecting} add computational overhead to the inference pipeline. To quantify this, we measure the clock time required for evaluation on the entire test set for each baseline, as shown in Figure \ref{fig:editing-memorization} (right). Our proposed approach stands out as more computationally efficient since it does not require any interference during inference. \footnote{We also add the cost of collecting neuron activations to calculate importance scores and pruning masks in the clock time. Since we consider $N=10$, the cost of collecting activations is very small.}

Along with this, we report the FID on the COCO30k dataset to check whether the model's generalization capabilities have been affected by the pruning. 
Figure \ref{fig:editing-memorization} (right) demonstrates that pruned models not only mitigate memorization but also retain their general image generation capabilities as evidenced by the low FID on the COCO30k dataset comparable to the no-mitigation baseline.

\subsection{An Intriguing Discovery -- Memorization Resides within a Potentially Unique Compact Subspace in Pre-Trained Models}

\begin{figure}[t]
    \centering
   \begin{subfigure}[b]{0.45\textwidth}
   \centering
       \includegraphics[width=0.9\textwidth]{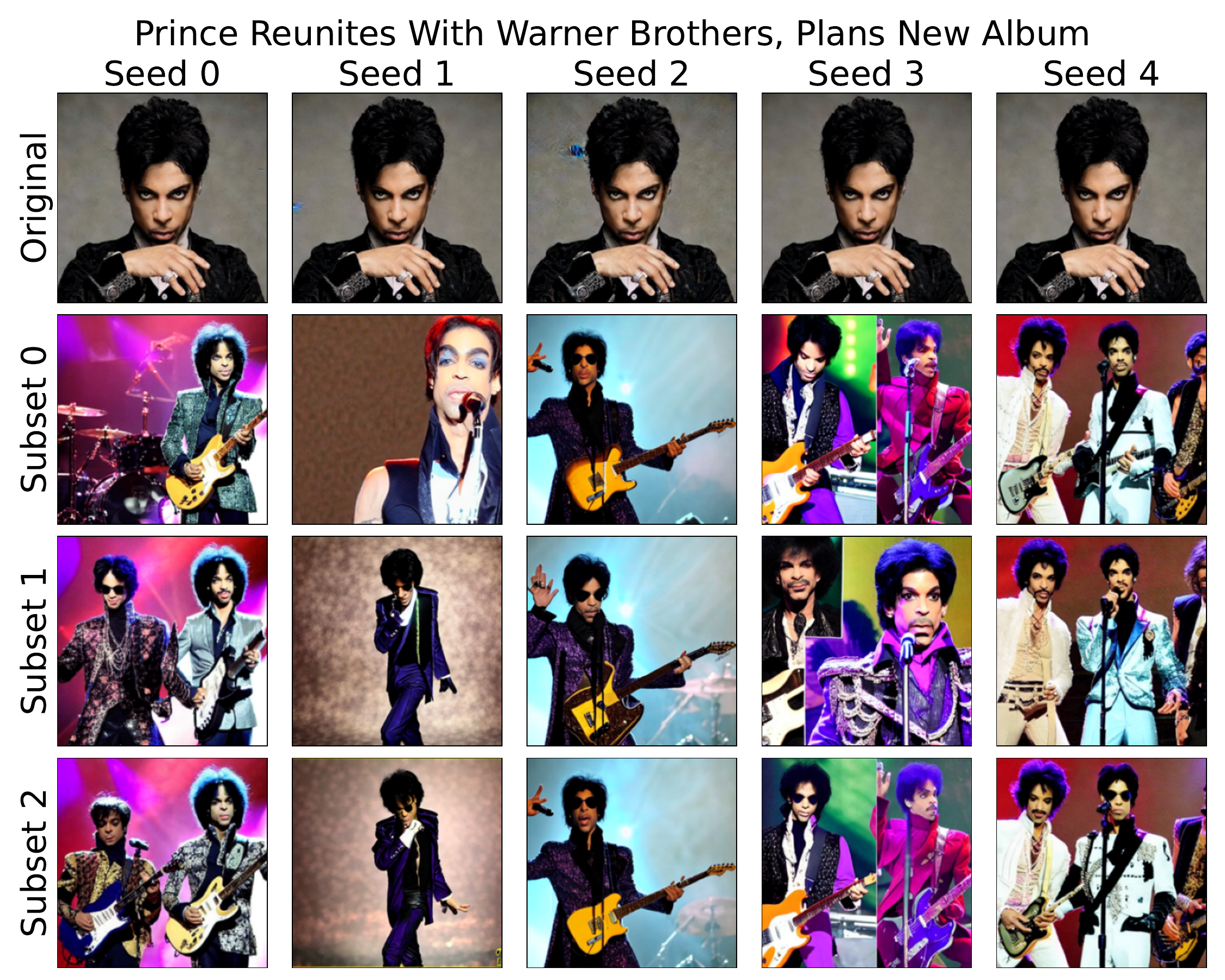}
   \end{subfigure}
   \begin{subfigure}[b]{0.45\textwidth}
   \centering
       \includegraphics[width=0.9\textwidth]{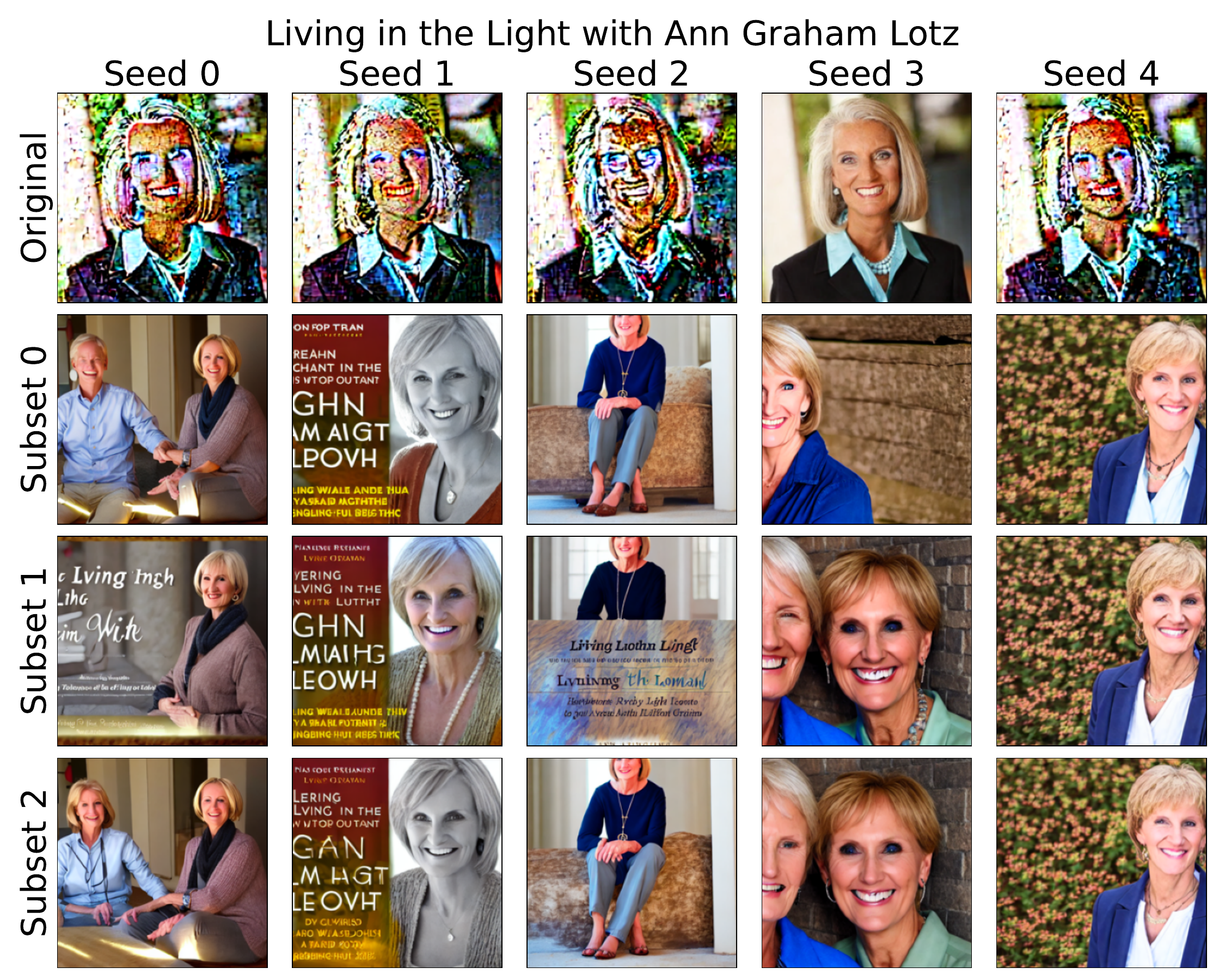}
   \end{subfigure}
   \caption{The initial row displays images generated by the pre-trained model, while subsequent rows depict images generated by different pruned models. Notably, despite sharing the same seed, different pruned models yield semantically similar images. This striking observation reveals that memorization resides in a potentially unique space in pre-trained diffusion models. {More qualitative results are presented in the appendix in Section \ref{sec:appendix}.}}
   \label{fig:grid}
\end{figure}

For text-to-image generation models, memorization is often characterized by overfitting to both the input prompt and a specific denoising trajectory. This manifests in generated images closely mirroring those in the training set, with minimal semantic variation across different initializations. Thus, effectively addressing memorization should result in output images that are (1) significantly different from the ones in the training set and (2) exhibit variability with diverse initialization. We demonstrate the former by evaluating pruned models on memorized prompts in Figure \ref{fig:editing-memorization}, showing that pruned models mitigate memorization. Furthermore, in the subsequent section, we demonstrate that extraction attacks on pruned models fail to retrieve training set images, indicating that our method prevents the close replication of training images. We demonstrate (2) in Figure \ref{fig:grid}, which shows variability in generated images with different initialization.

A notable observation from our results in Figure \ref{fig:editing-memorization} is that pruned models derived from different subsets exhibit consistent efficiency in mitigating memorization. Additionally, there is a significant overlap among the memorized neurons as seen in Figure \ref{fig:iou-different-subsets}. This points to a compelling conclusion -  \textit{Memorization resides within a potentially unique and compact subspace in pre-trained diffusion models.} 

Figure \ref{fig:grid} further bolsters our conclusion by illustrating that images generated from distinct pruned models, despite sharing the same seed, exhibit semantic similarity, implying significant overlap in pruned regions across these models.






\subsection{Pruned Models Effectively Resist Extraction Attacks}
The previous section evaluated the ability of our approach to alleviate memorization using a pre-identified set of memorized prompts. We now go beyond this analysis and conduct fine-tuning that leads to new memorizations, before showing that we can identify and remove those new memorizations with our pruning-based approach. More specifically we use the extraction attack from \cite{carlini2023extracting} to find the memorized images. After using our method, the attack does not identify memorized examples, indicating we mitigate the memorization.

\keypoint{Extraction attack.}
The attack from \citet{carlini2023extracting} consists of two main parts: 1) Generation of many image samples for each prompt using the generative model, and 2) Identification of memorized images using membership inference. \citet{carlini2023extracting} perform membership inference by constructing a graph of similar samples and finding cliques, which are groups of samples where each item is similar to all other items in the group. If a clique is sufficiently large, the samples within the clique are likely similar to the associated image, which means this image is likely memorized. We follow \citet{carlini2023extracting} in measuring similarity as the maximum $\ell_2$ distance across corresponding tiles of the two compared images. For our experiments we generate 50 samples for each prompt, use threshold of 50.0 for measuring similarity via the modified $\ell_2$ distance \citep{carlini2023extracting} and use minimum clique size of 3 when searching for potentially memorized images. The value of the threshold was selected so that visually similar images can be identified as similar.

\keypoint{Experimental setup.} We fine-tune Stable Diffusion v1.5 on Imagenette \citep{Howard_Imagenette_2019}
for 15,000 iterations with a batch size of 4. We randomly duplicate 100 images 50 times {in order to easily identify the potentially memorized images in the training set. } We then apply our memorization identification and pruning method to the fine-tuned model to compare the memorization before and after fine-tuning.

We present the results in Table~\ref{tab:ft}. The extraction attack identifies 9 examples out of the 100 to be potentially memorized, from which 8 are actually similar to the images in the set of duplicated images and hence are memorized. This shows the models can indeed memorize duplicated images through fine-tuning. After applying our method to the fine-tuned model, we successfully reduce the memorization rate to 0\%, demonstrating its effectiveness efficacy.

\begin{table}[h]
\centering
\caption{Memorization rate before and after pruning of the fine-tuned model. We report the proportion of examples that the attack identifies as memorized, and from these how many are actually memorized. Our pruning effectively removes the memorized images.}
\label{tab:ft}
\begin{center}
\resizebox{0.6\linewidth}{!}{
\begin{tabular}{lccl}
\toprule
 & \textbf{Before Pruning (\%)}  & \textbf{After Pruning (\%)} \\ 
\midrule
Identified as Memorized &  9\%  &  0\%   \\
Actually Memorized &  8\%  &  0\%   \\
\bottomrule
\end{tabular}
}
\end{center}
\end{table}

\begin{figure}
    \centering
    \includegraphics[width=0.8\linewidth]{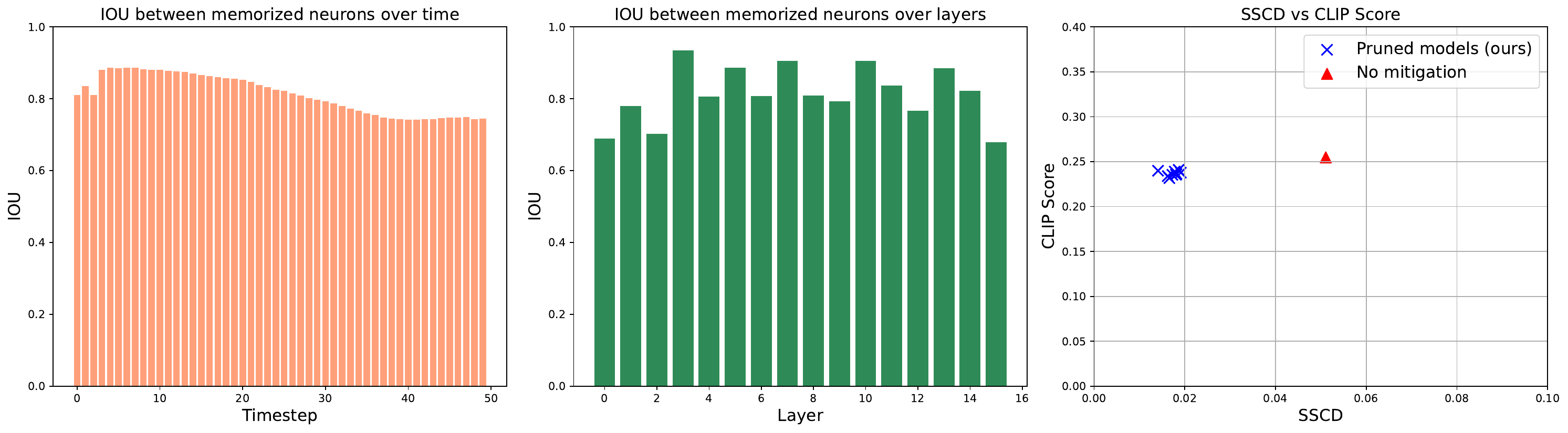}
    \caption{\textit{Left and Middle}: IOU between memorized neurons discovered from different subsets of memorized prompts is high, indicating localization of memorization. \textit{Right}: Memorization in SD2.0 can be mitigated with our proposed approach, indicating its generalizability across different models. }
    \label{fig:sd2}
\end{figure}




\subsection{Generalisation to Other Diffusion Models}

In the preceding sections, our focus was on Stable Diffusion 1.5. However, in this section, we extend our investigation to other diffusion models, specifically Stable Diffusion 2.0. Following a similar methodology as in previous sections, we apply our proposed approach to SD 2.0 and identify a collection of memorized neurons, as detailed in Section \ref{sec:mem-method}. Our visualizations in Figure \ref{fig:sd2} (left and middle) depict the similarities among memorized neurons, aligning with our earlier findings that distinct subsets of memorized prompts uncover highly similar sets of memorized neurons. Moreover, as illustrated in Figure \ref{fig:sd2}, we observe a decrease in SSCD, indicating that memorization can indeed be alleviated from pre-trained models through the pruning of memorized neurons. Therefore, our findings demonstrate that memorization is \textit{localized} to a specific compact subspace within the text-to-image generation model, and our proposed approach effectively identifies and mitigates it.  

\section{Limitations}
One limitation of our proposed approach is its reliance on a small set of memorized prompts as a starting point. While we demonstrate the ability to localize memorization with subsets as small as 10 prompts, certain inference-time mitigation techniques do not necessitate memorized prompts but instead introduce heuristics to identify memorization, potentially requiring access to a larger memorized dataset.

\section{Conclusions}
This study was inspired by safety-critical region identification in large language models (LLMs) and investigated critical neurons for the prompt memorization defect in pre-trained Diffusion Models (DMs). We followed a \emph{localize-and-prune} perspective. A recent SoTA weight pruning method, Wanda, is repurposed by employing its pruning strategy based on the collective effect of weights and input features, such that the important neurons in DMs for memorisation can be localized and then pruned. This is the first time the memorization of a DM can be mitigated in a training-free way. Various quantitative and qualitative evaluations demonstrated the strong efficacy of our method on memorization mitigation, outperforming the prior more sophisticated methods. Moreover, our pruned model is more robust to data extraction attacks, further showing its trustworthiness.

\newpage

\bibliographystyle{plainnat}
\bibliography{main}

\newpage
\section{Appendix}
\subsection{Qualitative visualizations of different pruned models}
\label{sec:appendix}
In this section, we present more examples similar to Figure \ref{fig:grid}. We present outputs of 10 pruned models for 5 different seeds and show that all pruned models output semantically similar images for a given seed, indicating that memorization is localized within a compact subspace in pre-trained models. 

\newpage

\begin{table}[]
    \centering
\begin{tabular}{cc}
\includegraphics[width=0.45\linewidth]{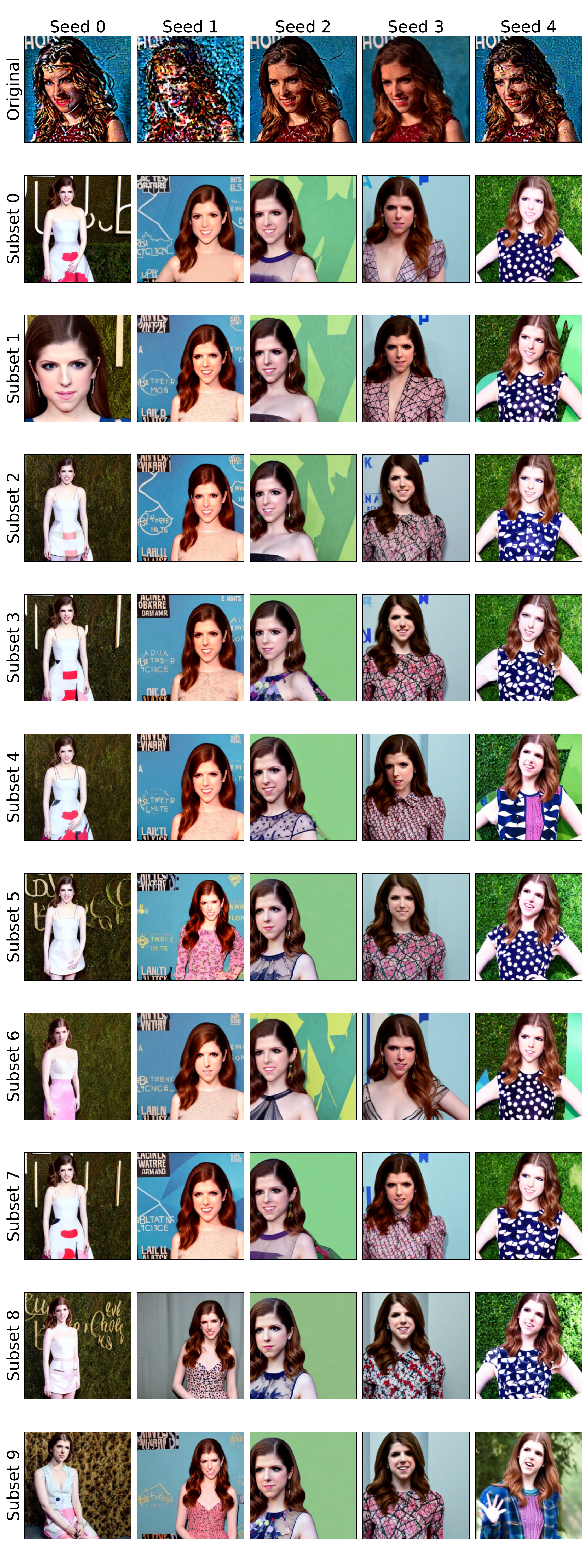} & \includegraphics[width=0.45\linewidth]{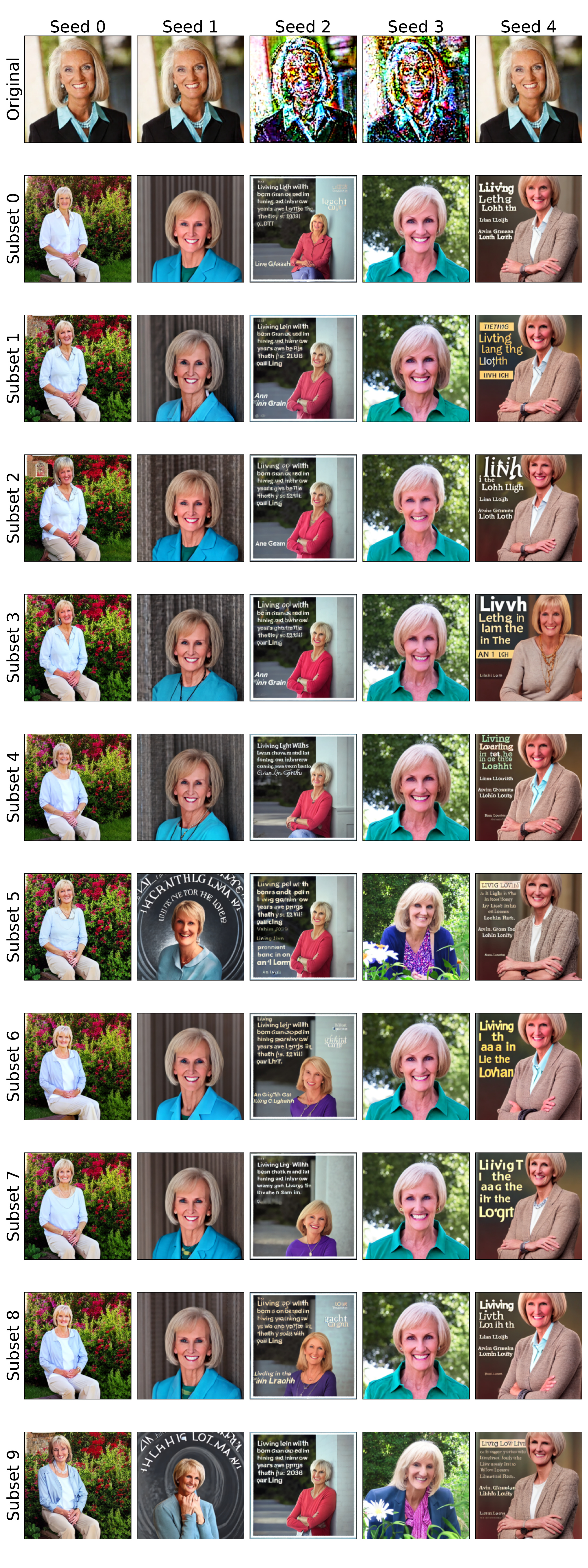}  \\
\end{tabular}
    \captionof{figure}{Qualitative results for memorised prompts. Left:  \textit{Anna Kendrick is Writing a Collection of Funny, Personal Essays}. Right: \textit{Living in the Light with Ann Graham Lotz} }
\label{fig:mem-grid-1}
\end{table}

\begin{table}[]
    \centering
\begin{tabular}{cc}
\includegraphics[width=0.45\linewidth]{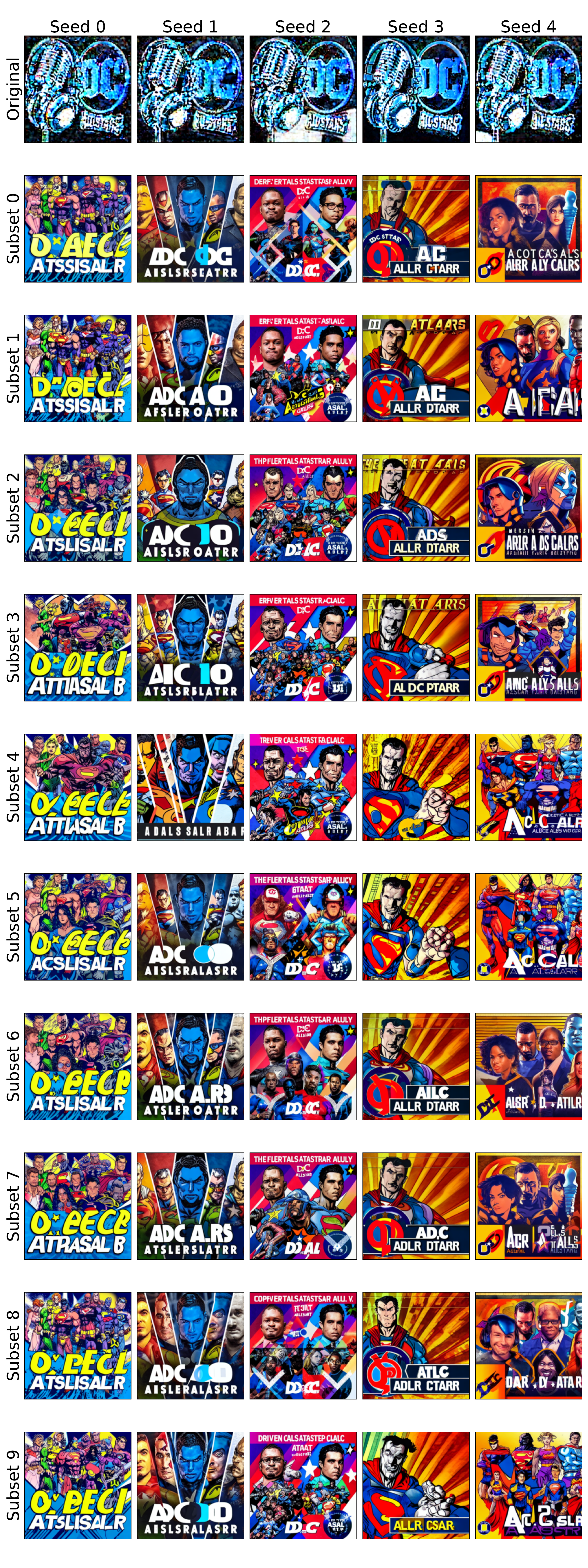} & \includegraphics[width=0.45\linewidth]{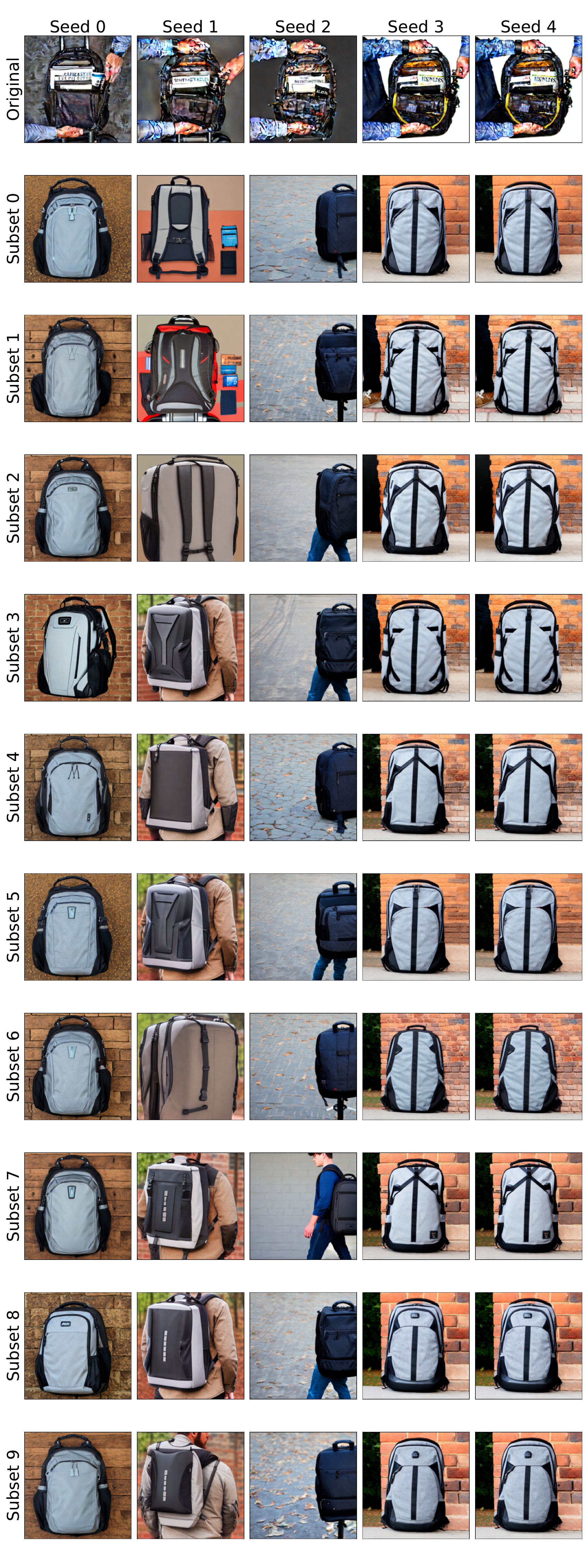}  \\
\end{tabular}
    \captionof{figure}{Qualitative results for memorised prompts. Left: \textit{DC All stars Podacst}. Right: \textit{Axle Laptop Backpack - View 81} }
\label{fig:mem-grid-1}
\end{table}

\begin{table}[]
    \centering
\begin{tabular}{cc}
\includegraphics[width=0.45\linewidth]{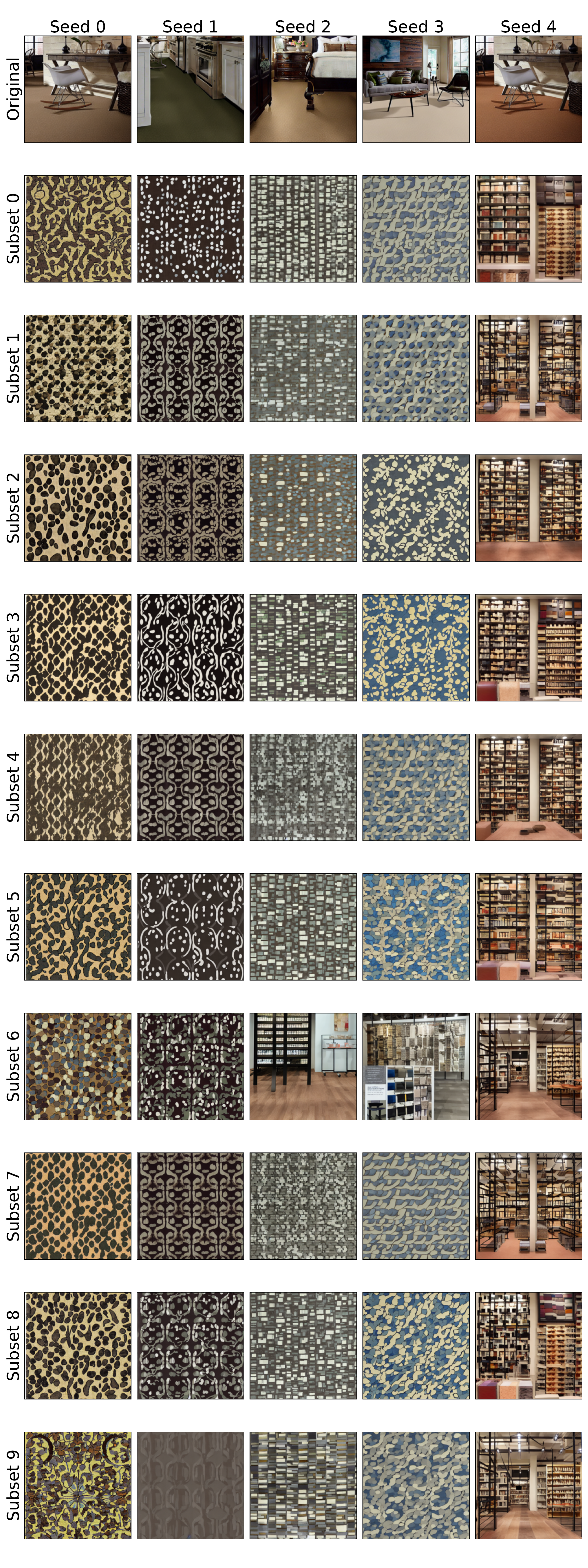} & \includegraphics[width=0.45\linewidth]{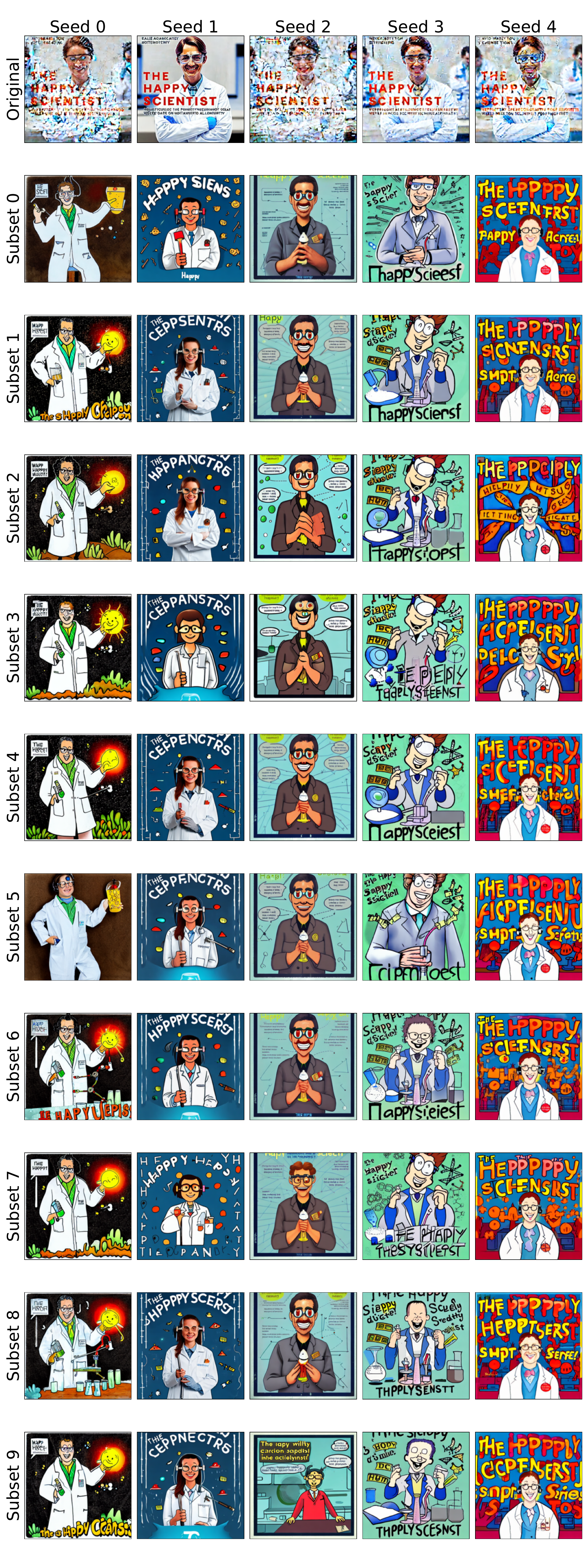}  \\
\end{tabular}
    \captionof{figure}{Qualitative results for memorised prompts. Left: \textit{Shaw Floors Shaw Design Center Different Times II 12 Silk 00104 5C494} Right: \textit{The Happy Scientist}}
\label{fig:mem-grid-1}
\end{table}



\end{document}